\documentclass[journal]{IEEEtran}
\usepackage{amsmath}
\usepackage{multirow}
\usepackage{graphicx}
\usepackage{hyperref}

\ifCLASSINFOpdf
\else
\fi
\begin{document}
%
\title{DFUNet: Convolutional Neural Networks for Diabetic Foot Ulcer Classification}
%
%
%

\author{Manu~Goyal, ~\IEEEmembership{Student Member,~IEEE,}
	~Neil~D.~Reeves,
	~Adrian~K.~Davison, ~\IEEEmembership{Member,~IEEE,}
	~Satyan~Rajbhandari,
	~Jennifer~Spragg, 
	~Moi~Hoon~Yap,~\IEEEmembership{Member,~IEEE}
	\thanks{M. Goyal and M. H. Yap are with the School of Computing, Mathematics and Digital Technology,
		Manchester Metropolitan University, John Dalton Building, M1 5GD, Manchester, UK.
		(e-mail: M.Yap@mmu.ac.uk)}
	\thanks{N.D. Reeves is with the School of Healthcare Science,
		Manchester Metropolitan University, John Dalton Building, M1 5GD, Manchester, UK.}
	\thanks{A.K. Davison is with the Centre of Imaging Sciences, University of Manchester, M13 9PL, Manchester, UK.}
	\thanks{S. Rajbhandari is with Lancashire Teaching Hospital, PR2 9HT, Preston, UK.}
	\thanks{J. Spragg is with Lancashire Care NHS Foundation Trust, PR5 6AW, Preston, UK.}}
\maketitle

\begin{abstract}
Globally, in 2016, one out of eleven adults suffered from Diabetes Mellitus. Diabetic Foot Ulcers (DFU) are a major complication of this disease, which if not managed properly can lead to amputation. Current clinical approaches to DFU treatment rely on patient and clinician vigilance, which has significant limitations such as the high cost involved in the diagnosis, treatment and lengthy care of the DFU. We collected an extensive dataset of foot images, which contain DFU from different patients. In this paper, we have proposed the use of traditional computer vision features for detecting foot ulcers among diabetic patients, which represent a cost-effective, remote and convenient healthcare solution. In this DFU classification problem, we assessed the two classes as normal skin (healthy skin) and abnormal skin (DFU). Furthermore, we used Convolutional Neural Networks for the first time in DFU classification. We have proposed a novel convolutional neural network architecture, DFUNet, with better feature extraction to identify the feature differences between healthy skin and the DFU. Using 10-fold cross-validation, DFUNet achieved an AUC score of 0.961. This outperformed both the machine learning and deep learning classifiers we have tested. Here we present the development of a novel and highly sensitive DFUNet for objectively detecting the presence of DFUs. This novel approach has the potential to deliver a paradigm shift in diabetic foot care.
\end{abstract}


%
\IEEEpeerreviewmaketitle

\section{Introduction}

\IEEEPARstart{D}{iabetes} Mellitus (DM)  commonly known as Diabetes, is a lifelong condition resulting from hyperglycemia (high blood sugar levels), which leads to major life-threatening complications such as cardiovascular diseases, kidney failure, blindness and lower limb amputation which is often preceded by Diabetic Foot Ulcers (DFU) \cite{wild2004global}. According to the global report on diabetes, in 2014, there are 422 million people living with DM compared to 108 million people in 1980. Among the adults that are over 18 years of age, the global pervalance has gone up from 4.7\% in 1980 to 8.5\% in 2014 \cite{world2016global}.  It is estimated by the end of 2035, the figure is expected to rise to 600 million people living with DM worldwide \cite{bakker20162015}. It is worth mentioning that about only 20\% of these people will be from developed countries and the rest will be from developing countries due to poor awareness and limited healthcare facilities \cite{boulton2005global}. There is about 15\%-25\% chance that a diabetic patient will eventually develop DFU and if proper care is not taken, that may result in lower limb amputation \cite{aguiree2013idf}.  Every year, more than 1 million patients suffering from diabetes lose part of their leg due to the failure to recognize and treat DFU appropriately \cite{armstrong1998validation}. A Diabetic patient with a 'high risk' foot needs periodic check-ups of doctors, continuous expensive medication, and hygienic personal care to avoid the further consequences as discussed earlier. Hence, it causes a great financial burden on the patients and their family, especially in developing countries where the cost of treating this disease can be equivalent to 5.7 years of annual income \cite{cavanagh2012cost}. 

In current clinical practices, the evaluation of DFU comprises of various important tasks in early diagnosis, keeping track of development and number of lengthy actions taken in the treatment and management of DFU for each particular case: 1) the medical history of the patient is evaluated; 2) a wound or diabetic foot specialist examines the DFU thoroughly; 3) additional tests like CT scans, MRI, X-Ray may be useful to help develop a treatment plan. The patient with DFU generally have a problem of a swollen leg, although it can be itchy and painful depending on each case. Usually, the DFU have irregular structures and uncertain outer boundaries. The visual appearance of DFU and its surrounding skin depending upon the various stages i.e. redness, callus formation, blisters, significant tissues types like granulation, slough, bleeding, scaly skin. Hence, the ulcer evaluation with the help of computer vision algorithms would be based on the exact assessment of these visual signs as color descriptors and texture features.

The major challenges that are involved with this classification task are as follows: 1) large time in collection and expert labelling of the DFU images 2) high inter-class similarity between the normal (healthy skin) and abnormal classes (DFU) and intraclass variations depending upon the classification of DFU \cite{lavery1996classification}, lighting conditions and patient's ethnicity. In this work, we have tested a number of Conventional Machine Learning (CML) methods and Convolutional Neural Networks (CNNs) for the classification of ulcer and non-ulcer. Then, we propose and design a novel fast CNN architecture, named as Diabetic Foot Ulcer Network (DFUNet), which outperformed GoogLeNet \cite{szegedy2015going} and AlexNet \cite{krizhevsky2012imagenet} in terms of \textit{accuracy} and \textit{sensitivity}.

\section{Related Work}
The proliferation of information and communication technologies present both challenges and opportunities in terms of the development of new age healthcare systems. There are a number of telemedicine systems that are currently being developed a) to improve the current healthcare systems and also, decrease the cost of medical facilities; b) to improve the reach of medical facilities i.e. frequent remote assessment of patients with the help of communication devices; c) to provide the automated solutions to deal with the shortage of expert medical professionals for these chronic diseases \cite{franc2011telemedicine}. Over the years, researchers and doctors have developed key telemedicine systems to monitor diabetes \cite{el2013systematic}. However, there are very few intelligent systems developed for assessment of diabetic foot pathologies which can be categorized into non-automated and automated telemedicine systems.

\subsection{Telemedicine Systems for DFU}
With the rapid growth in mobile telecommunications, remote communication is made possible with the help of standalone devices like smart-phones, laptops and Internet. Nowadays, a pocket size smart-phone with the advanced mobile operating system has the capability of a personal computer that can capture and send high-resolution pictures and also, audio and video communication with the help of advanced mobile internet like 4G. In the non-automated category, the common telemedicine systems based on these devices that are mostly set-up in the remote location for assessment of patients a) video conferencing \cite{clemensen2008treatment}; b) three-dimensional (3D) wound imaging \cite{bowling2011remote}; c) digital photography \cite{hazenberg2014assessment}; d) optical scanner \cite{foltynski2011monitoring}. However, there is still need of specialized medical professionals on the other side for completing the assessment of the patient. Though these systems provide promising results, but there is an urgent need of intelligent systems which can automatically detect the DFU pathologies remotely. 

The use of automated telemedicine systems for DFU is still in its infancy. Notably, Liu et al. \cite{liu2015automatic,van2014diagnostic} in 2015 developed an intelligent telemedicine system for detection of diabetic foot complications with the help of spectral imaging, infra-red thermal images and 3D surface reconstruction. However, to implement this system, there is a requirement of several expensive devices and specialist training to use these devices.  Wang et. al. \cite{wang2016area} have used an image capture box to capture image data and determined the area of DFU using cascaded two staged Support Vector Machine based classification. They proposed the use of a super-pixel technique for segmentation and extracted the number of features to perform two staged classification. Although this system reported a promising result, it has not been validated on a large dataset. In addition, the image capture box is very impractical for data collection as there is need for contact of the patient's feet and box surface which would not be allowed in a healthcare setting because of concerns regarding infection control. In other significant work, Manu et al. \cite{goyal2017fully} perform the segmentation of DFU and surrounding skin on the full foot images. 

Additionally,  computer methods based on manually engineered features or image processing approaches were implemented for tissue classification and segmentation of related skin lesion such as wound. The conventional machine learning for classification task was performed by extracting various features such as texture descriptors and color descriptors on small delineated patches of wound images, followed by machine learning algorithms to classify them into normal and abnormal skin patches \cite{wannous2011enhanced,kolesnik2005multi,kolesnik2006robust,papazoglou2010image,veredas2010binary}. As in many computer vision systems, the hand-crafted features are affected by lighting conditions and skin color depending upon the ethnicity group of the patient. In general, virtually all the skin lesions related to both wound and ulcer are now termed as wound. In medical perspective, both wound and ulcer are considered differently as wound are caused by an external problem whereas, ulcer are caused by an internal problem. Also, there are differences in appearance of the skin lesion of wound and ulcer, the cause (aetiology), the way the body responds (physiology) and disease processes (pathology) \cite{hermans2010wounds}. Hence, in this present study, only DFU are considered to determine how they are different from the normal healthy skin at the same place of appearance. 

\subsection{Computer Vision and Deep Learning}

In recent years, there has been a rapid development in the area of computer vision, especially towards the difficult and important issues like understanding images of different domains such as spectral, medical, object and face detection, multi-class and label classification \cite{zeiler2014visualizing,szegedy2015going, anthimopoulos2016lung,shin2016deep}. The conventional computer vision and machine learning algorithms were very limited in their ability to process the large image data, provide the representations of data with multiple levels of abstraction, and require a lot of manual tuning for each input image. Deep convolutional networks as a recent machine learning algorithm comes out as an important technique to solve these kinds of computer vision problems \cite{lecun2015deep,krizhevsky2012imagenet}. Deep convolutional networks obtain the multiple levels of representation methods by simple non-linear modules which transform the simple feature representation into the more advanced abstract representations for classification. Deep convolutional networks use images as input and start to learn features such as edges at specific directions and positions from the array of pixel values. At higher level, it combines these edges to learn more important abstract features such as components of desirable objects  and finally, these components are connected with each other to form final objects \cite{lecun2015deep}. 

Supervised learning is one of the most common forms of machine learning. It is very important for the training of the network as the system learn the classification tasks from a large collection of images that are labelled differently for each category. Without training, it is not possible for the machine to detect the desired category by giving the highest score of all categories \cite{simonyan2014very,long2015fully,tajbakhsh2016convolutional}. During the training stage, different images are processed by the machine to produce the output vector of scores for all categories for each image  and then, the error is measured in respect of output scores versus the expected score until the desirable score for each category is obtained. After training, a validation set of data or images is used to fine tune the hyper-parameters of networks like setting the weights for each layer and the number of convolutional and pooling layers. Lastly, the system is tested with real world test data without any expected outcome to check the performance of the system.  

The major contributions of this paper are as follows: 1) to the best of our knowledge, this is the first time, CNNs have been used to develop a fully automatic method to classify the DFU skin against the normal skin. 2) development of a novel CNN architecture called DFUNet, which is fine tuned to process the input data more effectively and efficiently than other comparative state-of-the-art CNNs architecture. The remainder of the paper is structured as follows. Section III describes the methodology that we used to design classifiers based on CML and CNNs and provides details of our proposed DFUNet. In Section IV, performance of various classifiers is tested with evaluation metrics like \textit{Sensitivity}, \textit{Specificity}, \textit{Precision}, \textit{F-Measure}, and \textit{Area Under the receiver operating characteristic Curve (AUC)}. In the Section V, the conclusion and future scope of our work are discussed.    

\section{Methodology}
This section describes the proposed dataset containing examples of DFU of various patients. This includes expert labelling of the different regions as normal and abnormal skin patches. In addition, the feature descriptors used in experiments are detailed, including for CML, the CNNs architecture of LeNet, AlexNet, and GoogLeNet. Finally, we propose our own CNN architecture, DFUNet, to improve the way DFU are classified.

\subsection{DFU Dataset}
The first challenge was to collect a dataset of standardized color images of DFU from various patients to train the various deep learning model. We utilized an extensive database of 292 images of patient's foot with DFU over the previous five years at the Lancashire Teaching Hospitals, obtaining ethical approval from all relevant bodies and patient’s written informed consent. Also, we collected 105 images of the healthy foot to get the more cases for normal healthy class. Approval was obtained from the NHS Research Ethics Committee to use these images for this research. These DFU images were captured with Nikon D3300. Whenever possible, the images were acquired with close-ups of the full foot with the distance of around 30-40 cm with the parallel orientation to the plane of an ulcer. The use of flash as the primary light source was avoided and instead, adequate room lights are used to get the consistent colors in images. To ensure the close range focus and avoiding the blurriness in images from the close distance, a Nikon AF-S DX Micro NIKKOR 40mm f/2.8G lens was used. We also included another test case that is captured by IPad with the help of FootSnap application to show the robustness of algorithms over heterogeneous capture setup \cite{yap2015computer}. It consists of 20 abnormal skin patches and 32 normal skin patches in this heterogeneous test case.   

\subsection{Expert Labelling of Images}
\begin{figure*}
	\centering
	\includegraphics[scale=0.28]{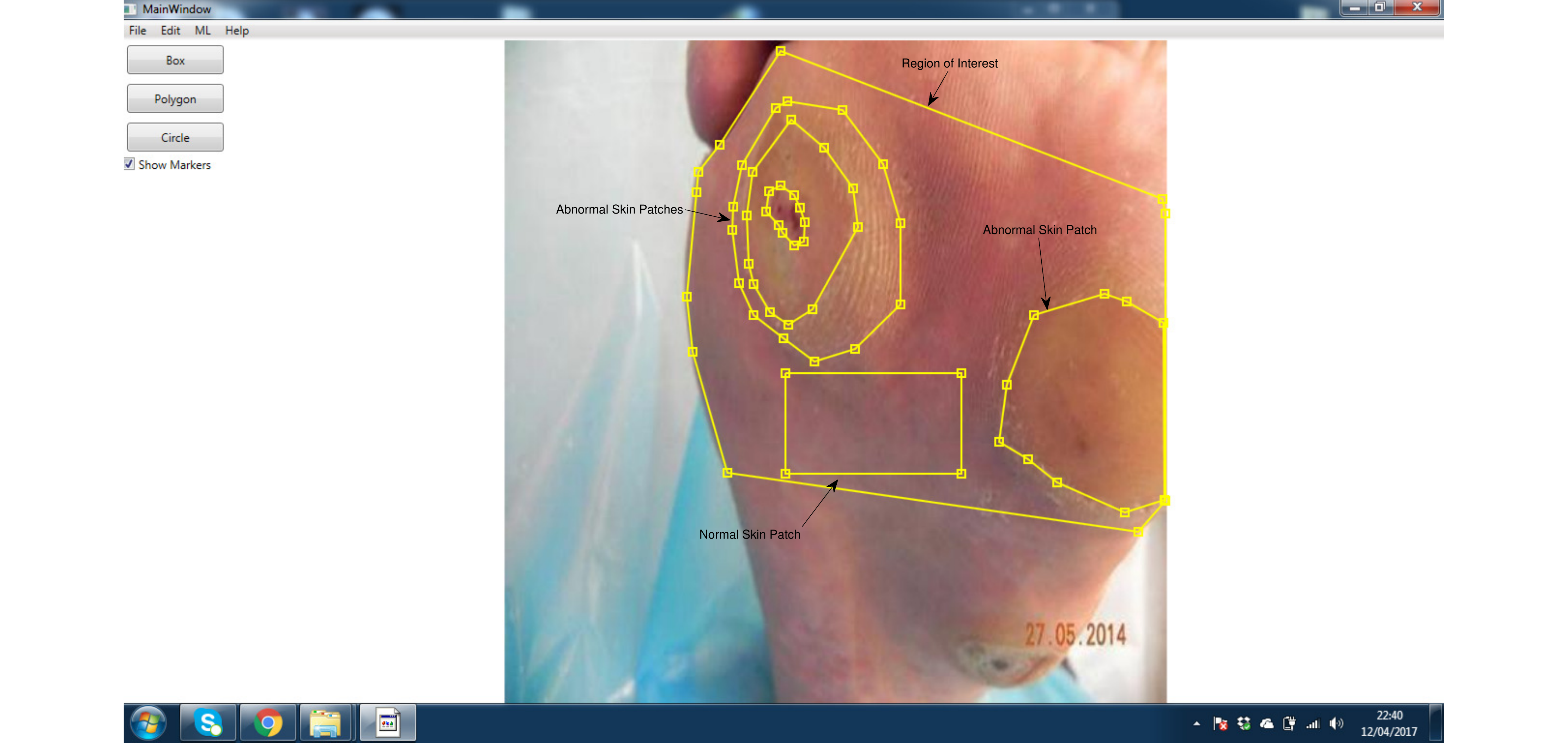}
	\caption{An example of delineating the different regions from the whole foot image}
	\label{fig:Annotator}
\end{figure*}

With the available annotator from Hewitt et al. \cite{hewitt2016manual}, for each full image of a foot with ulcers (as illustrated in Fig. 1), the medical experts delineated the Region Of Interest (ROI) which is an important region around the ulcer comprises of significant tissues of both normal and abnormal skin. The ground truth labels are delineated by medical professionals in the form of both normal and abnormal skin patches from the ROI region. In the collection of ground truth patches, the experts only collected both classes of patches from ROI region that helped with more robust classification of the patches rather than involving the whole foot as a region. For each delineated abnormal region, the ground truth of the type of the abnormality was labelled and exported to an Extensible Markup Language (XML) file. For the annotation on 397-foot images with both ulcer and non-ulcer, there is a total of 292 ROI (Only for the foot images with ulcers). From these annotations, we produce a total of 1679 skin patches with 641 normal and 1038 abnormal. Finally, we divided the dataset into training set of 1423 patches, validation set of 84 patches and testing set of 172 patches. The annotator tool which can delineate the image into different types of patches is shown in Fig. \ref{fig:Annotator}. 

\subsection{Data Augmentation of Training Patches}
Deep networks require a lot of training image data because of the enormous number of parameters, especially weights associated with convolutional layers needed to be tuned by learning algorithms. Hence, we used data augmentation to improve the performance by the deep learning methods. We used the combination of various image processing techniques like rotation, flipping, contrast enhancement, using different color space, and random scaling to perform data augmentation. The rotation is performed by rotating the image by angle of 90$^{\circ}$, 180$^{\circ}$, 270$^{\circ}$. Then, three types of flipping (horizontal flip, vertical flip and horizontal+vertical flip) performed on the original patches. The four color space that are used for data augmentation are \textit{Ycbcr}, \textit{NTSC}, \textit{HSV} and \textit{L*a*b}. In the contrast enhancement, we used the three functions called adjust image intensity value, enhanced contrast using histogram equalization, and contrast-limited adaptive histogram equalization. We produced the 2 times cropped patches with the help of random offset and random orientation from the original dataset of skin patches. With these techniques, we increase the number of training and validation patches by 15 times i.e. 21,345 patches for training and 1260 patches for validation.

\subsection{Pre-processing of Training Patches}
Since, we obtained the large number of training data with the help of data augmentation, it is very important to perform pre-processing on these patches. We used the zero-centre technique for pre-processing of these obtained patches, and then performed the normalization of every pixel.

\subsection{Conventional Machine Learning}
We investigate the use of human design features with CML on DFU and healthy skin classification. From our observation on the differences between DFU and healthy skin, the color and texture features descriptors were the visual cues for classification. For this 2-class classification problem, the sequential minimal optimization (SMO) \cite{platt199912} was selected as SVM based machine learning classifier.   

\subsubsection{Feature Descriptors}

We resize the patches of the whole dataset to 256$\times$256 to extract the uniform color and texture feature descriptors. The three color space that we have used: \textit{RGB}, \textit{HSV} and \textit{L*u*v}.

Local Binary Patterns (LBP) \cite{he1990texture} is one of the most popular texture descriptors for the classification. In our case, the LBP features are extracted to recognize the sudden change in texture in an abnormal region of the foot for detection of DFU.

Histogram of Oriented Gradients (HOG) \cite{dalal2005histograms} is a manually designed feature which converts the pixel based representation into the gradient based. In the context of this classification, HOG can be useful in terms of image gradients at an abnormal location in an image which gives you the intensity change in that location. As the gradient is a vector quantity, it has both the magnitude and direction.    

\subsection{Convolutional Neural Networks}
For comparison with the traditional features, deep learning, specifically convolutional neural networks, have been used to classify between healthy foot skin and skin with diabetic ulcerations. The first architecture we used was LeNet~\cite{lecun1995convolutional} running for 60 epochs, a learning rate of 0.01 with a step-down policy and step size of 33\%, and gamma is set to 0.1. This network was originally used for recognizing digits and zip codes. These simple structures are easily recognized, even in hand-written datasets such as MNIST \cite{lecun1998mnist}. 

\begin{figure}
	\centering
	\includegraphics[scale=0.8]{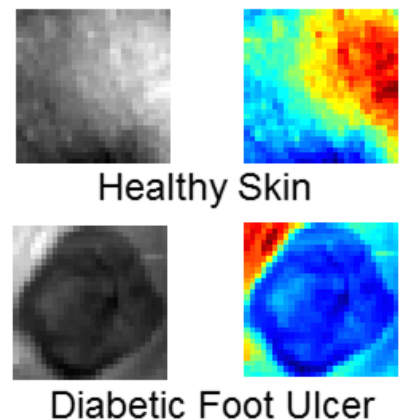}
	\caption{An example of the raw input (left) from the DFU dataset and the first activation from the LeNet architecture (right).}
	\label{fig:ulcerActivation}
\end{figure}

\begin{figure}
	\centering
	\includegraphics[scale=0.7]{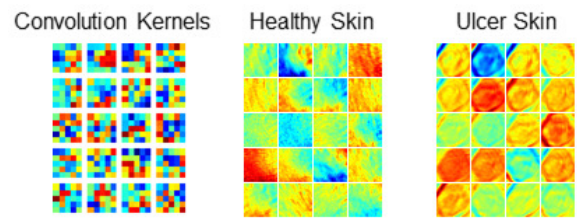}
	\caption{The output of healthy and diabetic ulcer skin from the first convolution layer of LeNet highlight discriminative features.}
	\label{fig:conv1output}
\end{figure}

Diabetic ulcers stand out on foot, as can be seen in Fig. \ref{fig:ulcerActivation} from an example from the diabetic ulceration dataset. Using LeNet represents these structures much better than traditional features, even on a relatively small training set of 1423 patches and validation of 84 patches.
The input was 28$\times$28 patches of skin in grayscale split into abnormal and normal skin samples. At the first convolution layer shown in Fig. \ref{fig:conv1output}, the kernels and activations already show the effectiveness of CNNs when highlighting important features.

We used the Caffe \cite{jia2014caffe} framework to implement LeNet \cite{lecun1995convolutional}, and used the Adaptive Moment Estimation (Adam) \cite{kingma2014adam} method for stochastic optimisation. This solver combines the advantages found in AdaGrad \cite{duchi2011adaptive}, which works well with sparse gradients, and RMSProp \cite{tieleman2012lecture}, which works well in an online setting. Adam is intended for large datasets and variability in parameters, however, the results in Table~\ref{tab:tradFeats} show that smaller datasets work just as effectively.

We also used popular CNN model AlexNet for classification of abnormal (DFU) and normal (healthy skin) classes. This network was originally used for classification of 1000 different objects of classes on ImageNet dataset. It emerged as winner of ImageNet ILSVRC-2012  competition in classification category by achieving 99\% confidence. There are few adjustments made in original network to work well for our 2-class classification problem. Also, a pre-trained model was used for better convergence of weights to achieve better results \cite{krizhevsky2012imagenet}. To train the model on Caffe framework, we used the same parameters as in LeNet i.e. 60 epochs, a learning rate of 0.01, and gamma of 0.1.

Another state-of-the-art CNN architecture that we used is GoogLeNet \cite{szegedy2015going}, a 22 layers deep network,  with similar experimental setting as of LeNet and AlexNet. Szegedy et al. \cite{szegedy2015going} introduced a new module called inception to GoogLenet. This acts as a multiple convolution filter inputs, that are processed on the same input and also does pooling at the same time. All the outcomes are then merged into single feature layer. This layer allows the model to take advantage of multi-level feature extraction from each input. Again, a transfer learning approach using pre-trained models to improve the performance.  

\subsection{Proposed Method - Diabetic Foot Ulcer Network}
To improve the extraction of important features for DFU classification, we propose a new Diabetic Foot Ulcer Network (DFUNet) architecture which is combination of important aspect of CNNs architecture - depth and parallel convolution layer. DFUNet combines two types of convolutional layers i.e. traditional convolution layers at the starting of the network which use single convolutional filter followed by parallel convolutional layers, which use multiple convolutional layers for extraction of multiple-features from the same input. Detecting changes in healthy skin is a clear computer vision problem similar to malignant skin lesions, so the DFUNet is designed around convolutions to finding discriminative features for learning.

\begin{figure*}
	\centering
	\includegraphics[scale=0.35]{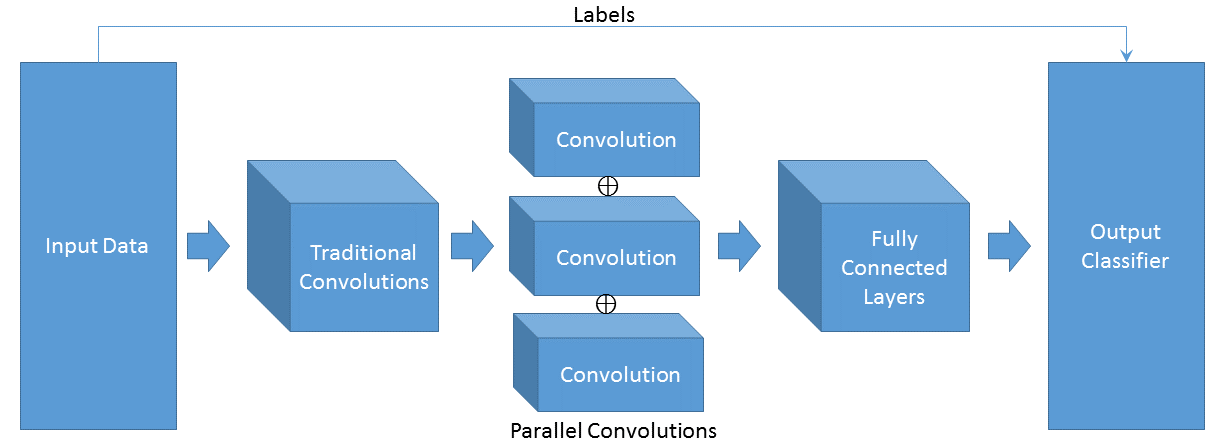}
	\caption{An overview of the proposed DFUNet architecture.}
	\label{fig:DFUNet}
\end{figure*}

Healthy skin tends to exhibit smooth textures and DFU have many distinct features including large edges, strong changes in intensity or color and quick changes between surrounding healthy skin and the DFU itself. DFUNet, summarised in Fig. \ref{fig:DFUNet}, is split into three main sections: the initialisation layers inspired by GoogLeNet, parallel convolution layers to discriminate the DFU more effectively than previous network layers and lastly, both fully-connected layers and a softmax-based output classifier. The detailed layers of the general DFUNet architecture are provided in the Table \ref{tab:networkarch}.
\begin{table*}[]
	\centering
	\small\addtolength{\tabcolsep}{-2pt}
	\renewcommand{\arraystretch}{1.5}
	\caption{Network Architecture of DFUNet. Conv. refers to convolutional layer, Max-pool. refers to Max-Pooling layers}
	\label{tab:networkarch}
	\scalebox{0.72}{
		\begin{tabular}{cccccccc}
			\hline
			Layer no. & Layer type    & Filter size                       & Stride     & No. of filters          & FC units & Input                    & Output                   \\ \hline\hline
			Layer 1   & Conv.         & 7$\times$7                        & 2$\times$2 & 64                      & -        & 3$\times$224$\times$224  & 64$\times$112$\times$112 \\
			Layer 2   & Max-pool.     & 3$\times$3                        & 2$\times$2 & -                       & -        & 64$\times$112$\times$112 & 64$\times$56$\times$56   \\
			Layer 3   & Conv.         & 1$\times$1                        & 1$\times$1 & 64                      & -        & 64$\times$56$\times$56   & 64$\times$56$\times$56   \\
			Layer 4   & Conv.         & 3$\times$3                        & 1$\times$1 & 192                     & -        & 64$\times$56$\times$56   & 192$\times$56$\times$56  \\
			Layer 5   & Max-pool.     & 3$\times$3                        & 2$\times$2 & -                       & -        & 192$\times$56$\times$56  & 192$\times$28$\times$28  \\
			Layer 6   & Parallel conv. & 1$\times$1,3$\times$3,5$\times$5 & 1$\times$1 & 32$\oplus$64$\oplus$128 & -        & 192$\times$28$\times$28  & 224$\times$28$\times$28  \\
			Layer 7   & Max-pool.     & 3$\times$3                        & 2$\times$2 & -                       & -        & 224$\times$28$\times$28  & 224$\times$14$\times$14  \\
			Layer 8   & Parallel conv. & 1$\times$1,3$\times$3,5$\times$5 & 1$\times$1 & 32$\oplus$64$\oplus$128 & -        & 224$\times$14$\times$14  & 224$\times$14$\times$14  \\
			Layer 9   & Parallel conv. & 1$\times$1,3$\times$3,5$\times$5 & 1$\times$1 & 32$\oplus$64$\oplus$128 & -        & 224$\times$14$\times$14  & 224$\times$14$\times$14  \\
			Layer 10  & Max-pool.     & 3$\times$3                        & 2$\times$2 & -                       & -        & 224$\times$14$\times$14  & 224$\times$7$\times$7    \\
			Layer 11  & Parallel conv. & 1$\times$1,3$\times$3,5$\times$5 & 1$\times$1 & 32$\oplus$64$\oplus$128 & -        & 224$\times$7$\times$7    & 224$\times$7$\times$7    \\
			Layer 12  & Max-pool.     & 7$\times$7                        & 1$\times$1 & -                       & -        & 224$\times$7$\times$7    & 224$\times$1$\times$1    \\
			Layer 13  & Fully conn.   & -                                 & -          & -                       & 1000       &                          &             \\
			Layer 14  & Fully conn.   & -                                 & -          & -                       & No. of Classes        &                          &        \\   \hline                          
	\end{tabular}}
\end{table*}

The parameters used for training with DFUNet are 40 epochs, a batch size of 8, the Adam solver with a learning rate of 0.001. A step-down policy is used where the learning rate reduces with a step of 33\% and gamma is set to 0.1.

\subsubsection{Input Data}
\begin{figure}
	\centering
	\includegraphics[scale=0.4]{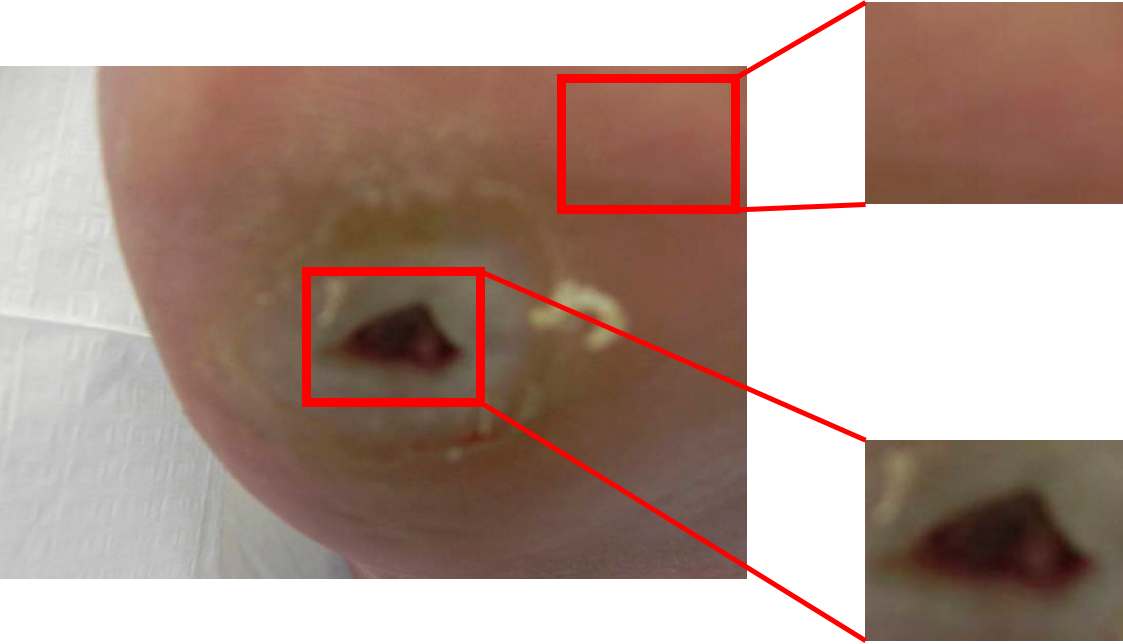}
	\caption{Healthy and ulcer patches taken from feet for training in the CNN.}
	\label{fig:dataInput}
\end{figure}

\begin{figure}
	\centering
	\includegraphics[scale=0.33]{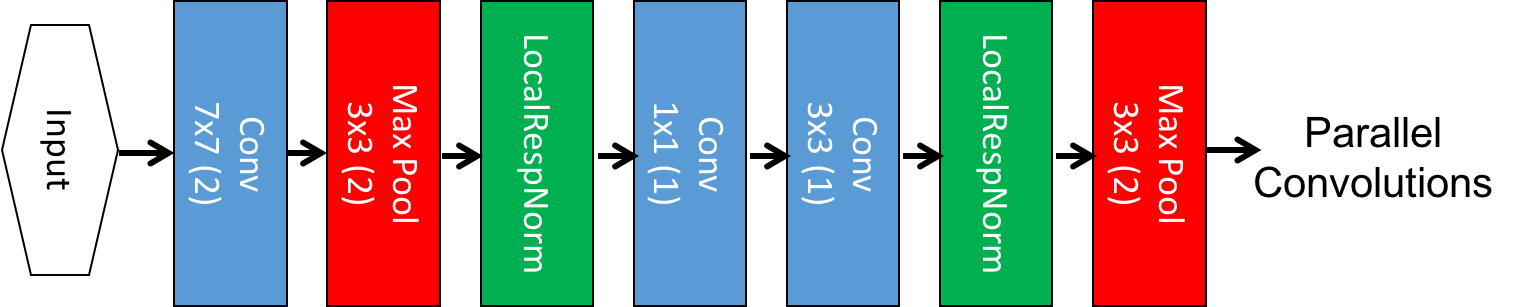}
	\caption{The initial input layers, similar to traditional CNNs, to prepare the data for the parallel convolution layers.}
	\label{fig:DFUNet_input}
\end{figure}

The DFU training and validation images are input as 256$\times$256 patches from areas of the feet containing DFU and healthy skin. An example of the regions of a foot cropped is shown in Fig. \ref{fig:dataInput}. We used the centre crop of size 224$\times$224 and mirror as data parameters. 
Inspired by the GoogLeNet \cite{szegedy2015going} input stem, the input to DFUNet, shown in Fig \ref{fig:DFUNet_input} begins by initial convolutions, pooling and normalisation layers in a traditional CNNs structure. Doing this step also ensures that the larger raw input image dimensionality is reduced before moving on to subsequent layers.

\subsubsection{Parallel Convolutions}
\begin{figure}
	\centering
	\includegraphics[scale=0.24]{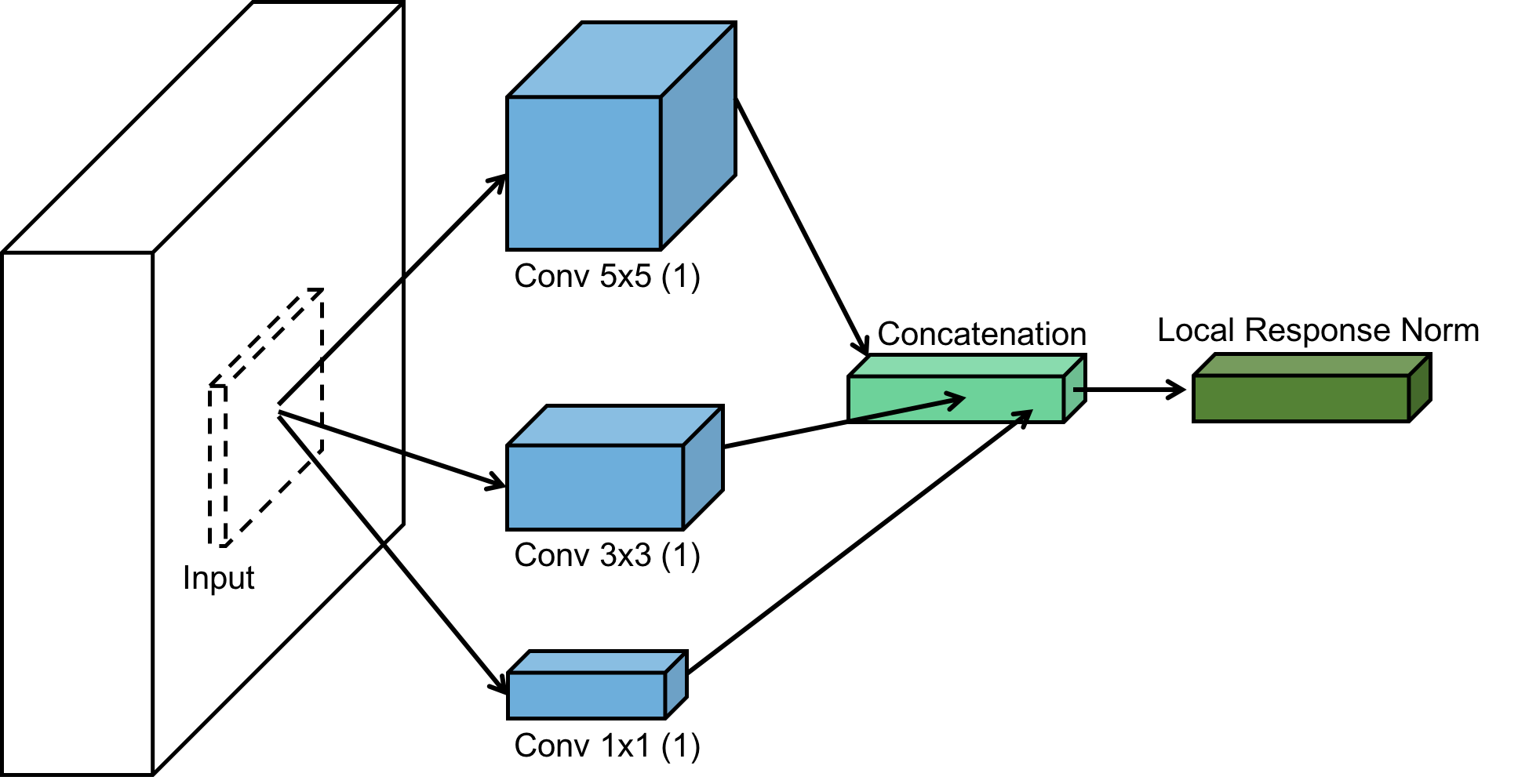}
	\caption{The structure of each parallel convolution layers.}
	\label{fig:DFUNet_conv}
\end{figure}

The traditional convolutional layers use only single type of convolutional filter popularly ranging from 1$\times$1 to 5$\times$5 on the input data. Each convolutional filter provides different feature extraction on the same input. The idea behind using the parallel convolutional layer is basically concatenation of multiple convolution filter inputs to allow the multiple-level feature extraction and cover more spread out clusters from same input. The design of the convolutions is weighted towards creating as discriminative features as possible to highlight any DFUs in an image. Three sizes of convolution kernels are used in the parallel convolutional layers of DFUNet throughout: 5$\times$5, 3$\times$3 and 1$\times$1. These are processed in parallel to each other and finally concatenated. The core of DFUNet is the four parallel convolutions and is shown in Fig. \ref{fig:DFUNet_conv}. The parallel convolutional layers are key innovation in methods appears to be in the architecture of the DFUNet. As this is the one of the most significant innovation, The DFUNet is experimented with different variants of these parallel sections to get the optimal architecture. There are total number of 5 variants of DFUNet is selected with different filter size that are experimented on the DFU dataset and the results are provided below in the Table \ref{tab:networkarcsh}.
	\begin{table}[]
	\centering
	\small\addtolength{\tabcolsep}{-2pt}
	\renewcommand{\arraystretch}{2.5}
	\caption{The performance measures of the various variants of DFUNet on DFU dataset. Conv. refers to convolutional layer and var. refers to variant}
	\label{tab:networkarcsh}
	\scalebox{0.6}{
		\begin{tabular}{cccccc}
			\hline
			Layers No.        & DFUNet Var. 1 & DFUNet Var. 2 & DFUNet Var. 3 & DFUNet Var. 4 & DFUNet Var. 5 \\ \hline\hline
			1st Parallel Conv. & 128$\oplus$256$\oplus$512      & 192$\oplus$256$\oplus$512    & 128$\oplus$128$\oplus$128      & 192$\oplus$192$\oplus$192      & 256$\oplus$256$\oplus$256      \\ 
			2nd Parallel Conv. & 128$\oplus$256$\oplus$512     & 192$\oplus$256$\oplus$512    & 128$\oplus$128$\oplus$128      & 256$\oplus$256$\oplus$256      & 256$\oplus$256$\oplus$256      \\ 
			3rd Parallel Conv. & 128$\oplus$256$\oplus$512      & 192$\oplus$256$\oplus$512    & 256$\oplus$256$\oplus$256      & 256$\oplus$256$\oplus$256   & 512$\oplus$512$\oplus$512   \\ 
			4th Parallel Conv. & 128$\oplus$256$\oplus$512      & 192$\oplus$256$\oplus$512    & 256$\oplus$256$\oplus$256      & 512$\oplus$512$\oplus$512   & 512$\oplus$512$\oplus$512  \\ \hline
	\end{tabular}}
\end{table}

\begin{figure}
	\centering
	\includegraphics[scale=0.6]{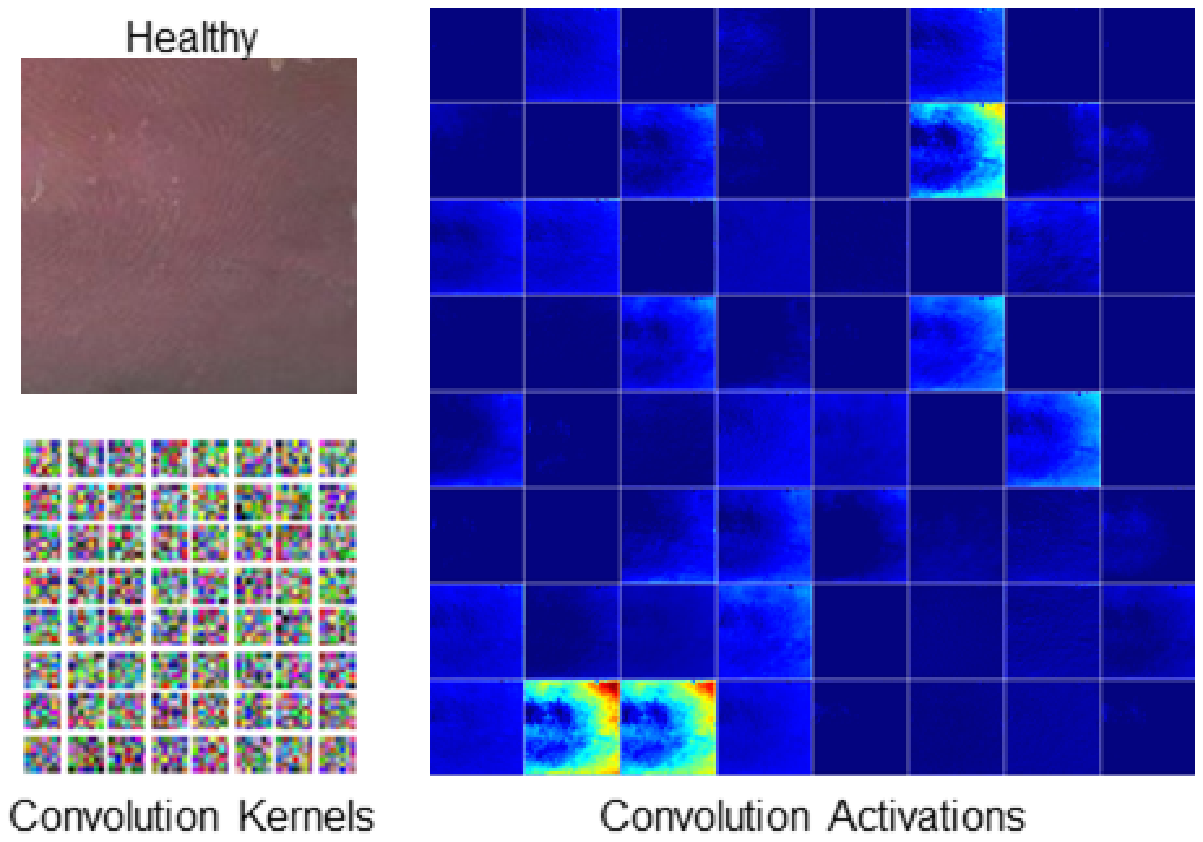}
	\caption{The healthy skin raw input, convolution kernels and convolution activations of DFUNet.}
	\label{fig:DFUNet_normConv}
\end{figure}

\begin{figure}
	\centering
	\includegraphics[scale=0.6]{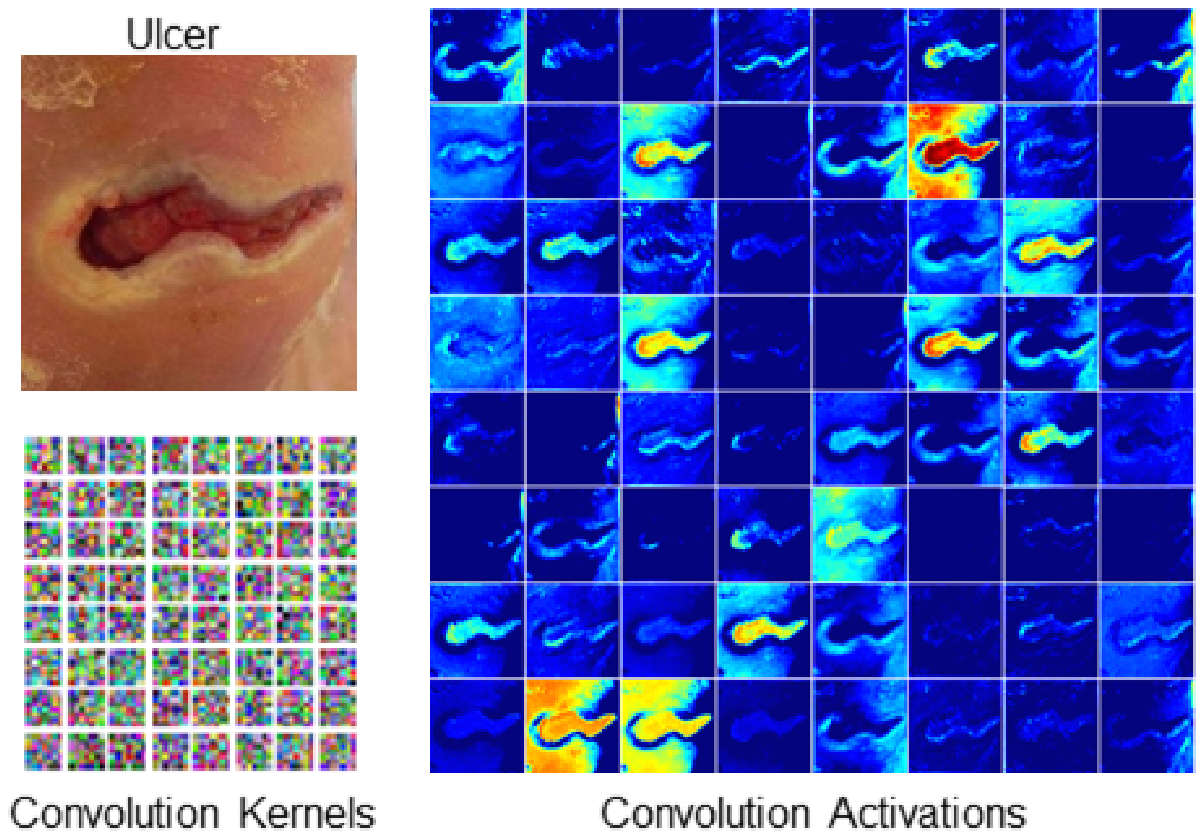}
	\caption{The diabetic ulcer skin raw input, convolution kernels and convolution activations of DFUNet.}
	\label{fig:DFUNet_abnormConv}
\end{figure}

Each convolution provides additional discriminative power. Lower activations are present in healthy skin samples shown in Fig. \ref{fig:DFUNet_normConv} due to the absence of skin abnormalities. Higher activations are present in skin with an ulcer as shown in Fig. \ref{fig:DFUNet_abnormConv} due to skin abnormality.

Each convolution layer uses a Rectified Linear Unit (ReLU) which is defined as
\begin{equation}
f(x) = \max(0, x)
\label{eq:relu}
\end{equation}
where the function thresholds the activations at zero. As we use a ReLU for each convolution, they include unbounded activations, so we use local response normalisation (LRN) to normalise these activations after each concatenation of convolutional layers. It is also proven helpful in avoiding the over-fitting problem faced by CNNs methods. Let, $a\tfrac{i}{x,y}$ be the source output of kernel \textit{i} applied at position \textit{(x,y)}. Then, regularized output $b\tfrac{i}{x,y}$ of kernel \textit{i} applied at position \textit{(x,y)} is computed by
\begin{equation}
b\tfrac{i}{x,y}=a\tfrac{i}{x,y} ( k+\alpha            \sum_{max(0,i-\frac{n}{2})}^{min(N-1,i+\frac{n}{2})}  (a\tfrac{j}{x,y})^{2 })^{\beta}
\label{eq:LRN}
\end{equation}
\noindent where N is total number of kernels, n is the size of the normalization neighbourhood and $\alpha$,$\beta$,\textit{k},(\textit{n}) are the hyper-parameters.

Further, to reduce dimensionality, a max pooling layer is included after the first and the third parallel convolutions.

\subsubsection{Fully Connected Layers and Output Classifier}
\begin{figure}
	\centering
	\includegraphics[scale=0.38]{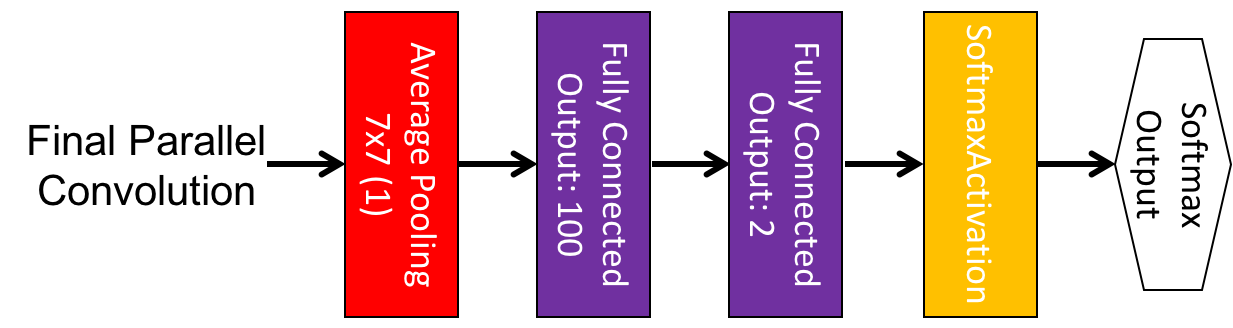}
	\caption{The final layers, including a softmax classifier, to predict normal skin and DFU.}
	\label{fig:outputClassifier}
\end{figure}

The final section is the softmax output of class probabilities and is a measure of how close the parameters are with respect to the ground truth labels of the training and validation data. The 2-class outputs of the DFU is healthy skin and DFU. It is formed from an average pooling layer followed by two fully connected (FC) layers with outputs of 100 for the first and 2 for the second. It is worth mentioning, the DFUNet is fine-tuned for the 2-class problem by using only outputs of 100 rather than 1000 in first FC layer and last FC layer is adjusted as 2. This fine-tuning helps in faster processing time in both training and testing phase of the DFUNet. The softmax function (cross-entropy regime) is the final layer and is defined as
\begin{equation}
f_j(z) = \frac{e^{z_j}}{\sum_k e^{z_k}}
\end{equation}
where $f_j$ is the $j$-th element of the vector of class scores $f$ and $z$ is a vector of arbitrary real-valued scores that are squashed to a vector of values between zero and one that sum to one. The loss function is defined so that having good predictions during training is equivalent to having a small loss. The output layers are summarized in Fig. \ref{fig:outputClassifier}. 

\section{Results and Discussion}
The DFU dataset was split into the 85\% training, 5\% validation and 10\% testing sets and we adopted the 10-fold cross-validation technique. Hence, for training and validation using the proposed DFUNet architecture, we used approximately 1423 patches (including 882 abnormal cases) and 84 patches (including 52 abnormal cases) respectively from the 397 original foot images. As mentioned previously, we used both CML models and CNNs models to do the classification task. LeNet was the only architecture that worked on 28$\times$28 gray scale patches rather than 256$\times$256 RGB images as input used by GoogLeNet, AlexNet, DFUNet and CML. It was included to show how the basic deep learning works on this new classification problem.      	 

With data augmentation technique, these patches are made 15 times  for both training and validation. But, when we used the data augmentation technique in our experiment, the final results are found to be the same with all the models. Hence, we did not include the data augmentation datasets in Table \ref{tab:dfuresult} and Table \ref{tab:tradFeats} as it did not improve the results. The main reasons behind the failure of data augmentation was overall performance metrics recorded without data augmentation was quite high and there was only small number of misclassification cases which were not corrected even with models trained with data augmentation.

In Table \ref{tab:tradFeats}, we report \textit{Sensitivity}, \textit{Specificity}, \textit{Precision}, \textit{Accuracy}, \textit{F-Measure} and \textit{Area under curve of ROC (AUC)} as our evaluation metrics. In medical imaging, \textit{Sensitivity} and \textit{Specificity} are considered reliable evaluation metrics for classifier completeness. 

\begin{equation}
Sensitivity= \frac {TP}{TP+FN}
\end{equation}

\begin{equation}
Specificity= \frac {TN}{FP+TN}
\end{equation}

\begin{equation}
Precision= \frac {TP}{TP+FP}
\end{equation}

\begin{equation}
Accuracy= \frac {TP+FN}{TP+TN+FP+FN}
\end{equation}

\begin{equation}
F-Measure= \frac {2*TP}{2*TP + FP + FN}
\end{equation}

\begin{figure}
	\centering
	\includegraphics[scale=0.45]{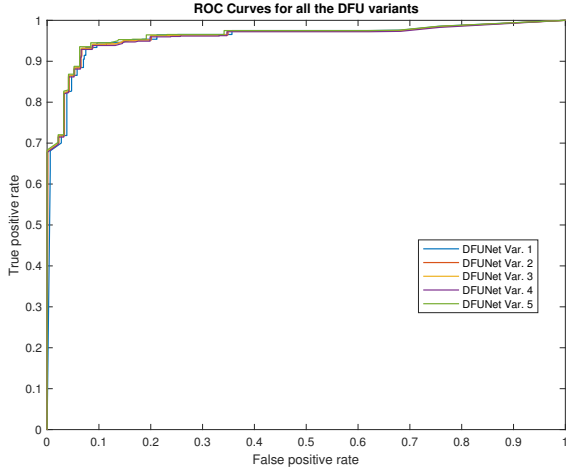}
	\caption{The ROC curve for all DFUNet models in which DFUNet var. 5 performed best with an AUC score of 0.961. Var. refers to variant.}
	\label{fig:Rocdfu}
\end{figure}

In Table \ref{tab:dfuresult}, we report the performance measures of various DFUNet variants with different parameters as explained in the architecture of DFUNet in previous section. There was not much gap in performances between all the models. But, overall, the DFUNet variant 5 performed best in every evaluation metrics except \textit{Precision} in which DFUNet variant 1 performed the best. It is clear that DFUNet variant 5 which uses the much larger filter sizes than other variants in last two parallel convolutional layers produced better results. Hence, with best results achieved by DFUNet variant, we used it as a proposed DFUNet to compare the performance with other traditional machine learning and deep learning models. ROC curve for all the variants is illustrated by Fig. \ref{fig:Rocdfu}. 

\begin{table}[]
	\centering
	\small\addtolength{\tabcolsep}{-2pt}
	\renewcommand{\arraystretch}{1.5}
	\caption{The performance measures of various variants of the DFUNet on DFU dataset without data augmentation}
	\label{tab:dfuresult}
	\scalebox{0.72}{
		\begin{tabular}{ccccccc}
			\hline
			& \textit{Sensitivity} & \textit{Specificity} & \textit{Precision} & \textit{Accuracy} & \textit{F-Measure} & \textit{AUC} \\ \hline\hline
			DFUNet Var. 1 & 0.923                & 0.910                  & \textbf{0.946}               & 0.918             & 0.934                &         0.957      \\ 
			DFUNet Var. 2 & 0.928                  & 0.905                 & 0.942                & 0.919               & 0.935                & 0.959          \\ 
			DFUNet Var. 3 & 0.928                  & 0.906                  & 0.942                & 0.921               & 0.935                &     0.960           \\ 
			DFUNet Var. 4 & 0.927                  & 0.900                  & 0.938                & 0.917               & 0.933                &      0.958          \\ 
			DFUNet Var. 5 & \textbf{0.934}                  & \textbf{0.911}                  & 0.945                & \textbf{0.925}               & \textbf{0.939}                &  \textbf{0.961}              \\ \hline
	\end{tabular}}
\end{table}

There are three CML models and three CNNs models used for classification. In CML, we used the combination of LBP, HOG and Colour descriptors (\textit{RGB}, \textit{HSV} and \textit{L*u*v}) as feature vectors and then, we trained an SMO for our classification problem. For each CNN, LeNet, AlexNet, GoogLeNet and our proposed DFUNet are the chosen architectures used for classification. Each classifier performed well for \textit{Sensitivity} with less than 1.4\% margin between the highest result (DFUNet) and the lowest result (LBP + HOG). There is a larger gap of 7.7\% in \textit{Specificity} for the CML models performance measure, with results ranging from 0.835 to 0.845.

For the CNNs approaches, LeNet achieved the lowest score of 0.81 for \textit{Specificity}, whereas the AlexNet, GoogLeNet and DFUNet performed best in this category, with 0.892, 0.912, and 0.908 respectively. AUC is considered to be a viable performance measure for the different machine learning approaches for classification, with DFUNet and GoogLeNet achieving 0.961 and 0.960 respectively. 

\begin{figure}
	\centering
	\includegraphics[scale=0.45]{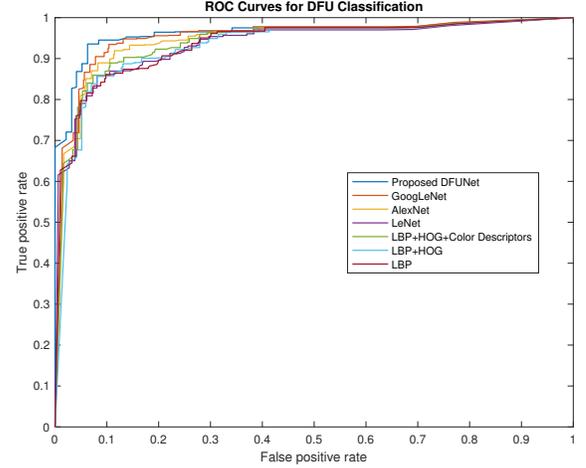}
	\caption{ROC curve for all the models including CML and CNNs mentioned in Table \ref{tab:tradFeats} in which our proposed DFUNet method achieved the best score.}
	\label{fig:roccurvewhole}
\end{figure}

\begin{table*}[]
	\centering
	\renewcommand{\arraystretch}{1.18}
	\caption{DFU classification results. Overall, our proposed DFUNet achieved the best results.}
	\label{tab:tradFeats}
	\begin{tabular}{cccccccc}
		\hline	& \textit{Sensitivity} & \textit{Specificity} & \textit{Precision} & \textit{Accuracy} & \textit{F-Measure} & \textit{AUC} \\ \hline\hline
		LBP                                   & 0.919       & 0.764       & 0.878     & 0.865    & 0.898     & 0.932    \\
		LBP + HOG                             & 0.881       & 0.841       & 0.906     & 0.866    & 0.893     & 0.931    \\
		LBP + HOG + Colour Descriptors         & 0.902       & 0.845       & 0.904     & 0.880    & 0.904     & 0.943    \\
		LeNet (CNN)\cite{lecun1995convolutional}                   			& 0.912       & 0.810       & 0.871     & 0.872    & 0.893    & 0.929    \\
		Alexnet (CNN)\cite{krizhevsky2012imagenet}              			& 0.895       & 0.886       & 0.933     & 0.893    & 0.914     & 0.950    \\
		GoogLeNet (CNN)\cite{szegedy2015going}               			& 0.905       & \textbf{0.912}       & \textbf{0.949}     & 0.907    & 0.927     & 0.960    \\
		Proposed DFUNet        			         & \textbf{0.934}                  & 0.911                  & 0.945                & \textbf{0.925}               & \textbf{0.939}                &  \textbf{0.961}   \\ \hline
	\end{tabular}
\end{table*}
Overall, we showed that using CNNs can outperform the more traditional CML features by a large margin. All CNN architectures achieved higher results than any of the CML results in most cases. GoogLeNet and DFUNet were the best performers for various evaluation metrics among all the classifiers. The ROC curve for all the models is demonstrated by the Fig. \ref{fig:roccurvewhole}. The details of AUC performance for each method is described in Table \ref{tab:dfuresults}.

\begin{table*}[]
	\centering
	\small\addtolength{\tabcolsep}{-2pt}
	\renewcommand{\arraystretch}{1.5}
	\caption{The performance measures of all methods on AUC curve}
	\label{tab:dfuresults}
	\scalebox{0.72}{
		\begin{tabular}{ccccccc}
		\hline	& \textit{AUC Score} & \textit{Standard Error of the Area} & \textit{Confidence interval of the AUC (95 percent)}   \\ \hline\hline
		LBP                                   & 0.9322       & 0.0061       & 0.9202 - 0.9443           \\
		LBP + HOG                             & 0.9308       & 0.0060       & 0.9190 - 0.9427           \\
		LBP + HOG + Colour Descriptors         & 0.9430       & 0.0054       & 0.9324 - 0.9537           \\
		LeNet (CNN)\cite{lecun1995convolutional}                   			& 0.9292       & 0.0060       & 0.9173 - 0.9412           \\
		Alexnet (CNN)\cite{krizhevsky2012imagenet}              			& 0.9504       & 0.0050     & 0.9405 - 0.9603            \\
		GoogLeNet (CNN)\cite{szegedy2015going}               			& 0.9604       &  0.0045     & 0.9514  -  0.9690            \\
		Proposed DFUNet        			         & \textbf{0.9608}                  & 0.0044                  & 0.9520 - 0.9695                                 \\ \hline
	\end{tabular}}
\end{table*}

We received better results than GoogLeNet on various evaluation metrics. The reason behind using the DFUNet rather than conventional CNNs architecture in particular GoogLeNet is to speed up the best results with the help of lesser layers i.e. 14 layers architecture compared to the 22 layers architecture of GoogLeNet and fine tuning the overall architecture of DFUNet according to the 2-class problem i.e. normal and abnormal skin patches. With the 10-fold cross validation, on the same machine configuration and input batchsize on Caffe framework, DFUNet took an average of 3 minutes 32 seconds where as GoogLeNet took average of 16 minutes 27 seconds to train a model with the same amount of training and validation data. For testing, DFUNet took an average of 49 seconds where as GoogLeNet took an average of 72 seconds to classify the same test data. Therefore, we demonstrated how reducing the number of layers using the bespoke architecture of DFUNet markedly reduced processing time, while also achieving a higher sensitivity and specificity with introduction of parallel convolution layers with increased number of filter input.

Our proposed DFUNet has highest performance measures in \textit{Sensitivity}, with a score of 0.934, \textit{F-measure} with 0.939 and \textit{AUC} with 0.962. Whereas, GoogLeNet has highest score in \textit{Specificity} and \textit{Precision} due to it's robust nature of being able to find more subtle changes using the inception architecture \cite{szegedy2015going}.

There is no evidence of an influence of factors such as lighting conditions and skin tone due to patient's ethnicity on DFU classification.  As ulcer and surrounding skin has quite distinctive texture and color features from the normal healthy skin irrespective of above mentioned factors. In our experiments, these factors result in very few misclassified instances in testing set when there is very high red skin tone as shown in Fig. \ref{fig:misclass}.

\subsection{Accurate and Inaccurate Cases of Classification by Proposed DFUNet}
\begin{figure*}
	\centering
	\includegraphics[scale=0.55]{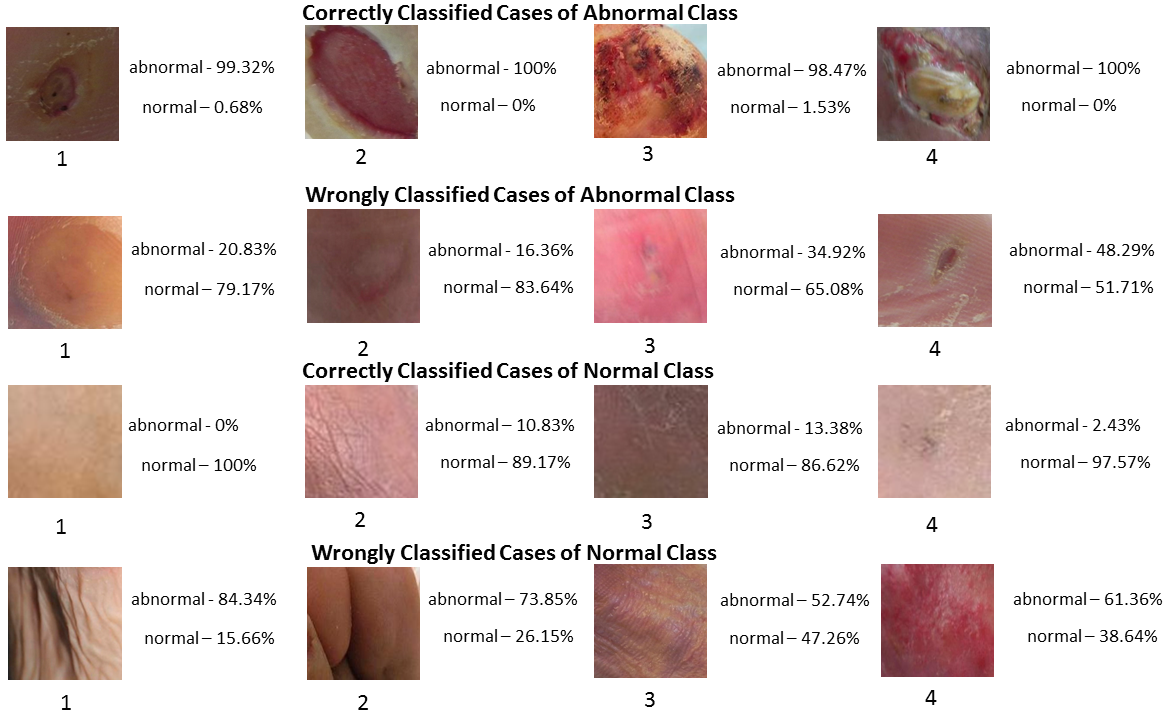}
	\caption{Correctly and wrongly classified cases for both abnormal and normal classes}
	\label{fig:misclass}
\end{figure*}

There are a few examples of correctly and incorrectly classified cases in both abnormal and normal classes as illustrated in Fig. \ref{fig:misclass}. The performance of DFUNet is quite accurate in correctly classifying most of the testing instances. DFUNet generally struggle to classify the pre-ulcer skin and usually detected it as normal with high percentage as illustrated by example 1 and 2 of misclassification cases of abnormal class in Fig. \ref{fig:misclass}. Also, DFU that are very small in size are misclassified as normal as shown by example 3 and 4 of misclassification cases of abnormal class Fig. \ref{fig:misclass}. In normal skin, the patches with toe, highly wrinkled skin,  and very high red tone skin are classified wrongly by the proposed method as illustrated by the examples of misclassified cases of normal classes in Fig. \ref{fig:misclass}.         

\section{Performance evaluation on Heterogeneous Test Case}
Since, DFU dataset is captured with the same DSLR camera as mentioned in above section. With computer vision techniques, it is preferable to have heterogeneous capture to form dataset. But, strict medical ethical approval does not allow to use different cameras to capture the pictures of DFU. Hence, we collected another heterogeneous dataset of standardized DFU images with the help of FootSnap application. These images are captured with the help of IPad camera. We tested our algorithm on this heterogeneous dataset and received good performance with \textit{Sensitivity} score of 0.929, \textit{F-measure} with 0.931, , \textit{Specificity} of 0.908, \textit{Precision} with 0.942 and \textit{AUC} with 0.950 score.      


\section{Performance Evaluation on Facial Skin Dataset}

Since, DFUNet performed well on the classification of DFU skin patches, to test the robustness of DFUNet on other skin lesion datasets, we run the experiment of 3-class classification of facial skin patches i.e. normal, spot and wrinkles. It is worth mentioning, there is no public skin lesion dataset available for research without prior written consent. In this derma dataset, we delineated the equal number of skin patches i.e. 110 patches for each class.  We used only two best performing CNN architectures in Table. \ref{tab:tradFeats} i.e. GoogLeNet and DFUNet for this experiment. With the same experimental settings, DFUNet outperforms GoogLeNet in each evaluation metrics for 10-fold cross-validation data as shown in Table \ref{tab:tradFs}. This is due to the deep learning models does not work well with smaller dataset even with full training \cite{tajbakhsh2016convolutional}. But, DFUNet uses larger filter sizes in the later parallel convolution layers to extract more multiple features which helps DFUNet outperforms GoogLeNet in this experiment.

\begin{table}[]	
	\centering
	\small\addtolength{\tabcolsep}{-2pt}
	\renewcommand{\arraystretch}{1.5}
	\caption{Facial Skin classification results. Overall, our proposed DFUNet achieved the best results.}
	\label{tab:tradFs}
	\scalebox{0.72}{
		\begin{tabular}{ccccccc}
			\hline	
			& \textit{Sensitivity} & \textit{Specificity} & \textit{Precision} & \textit{Accuracy} & \textit{F-Measure} & \textit{MCC} \\ \hline\hline
			GoogLenet       & 0.783       & 0.882       & 0.784     & 0.846    & 0.784     & 0.665 \\
			Proposed DFUNet & 0.867       & 0.930       & 0.867     & 0.907    & 0.867     & 0.796 \\\hline
	\end{tabular}}
\end{table} 

\section{Conclusion}
In this work, we trained various classifiers based on traditional machine learning algorithms, CNNs and proposed a new CNN architecture, DFUNet on DFU classification which discriminates the DFU skin from healthy skin. With high-performance measures in classification, DFUNet allows the accurate automated detection of DFU in foot images and make it an innovative technique for DFU evaluation and medical treatment. For the detection of DFU, it is very important to understand the difference between DFU and healthy skin to know the features differences between these two classes in computer vision perspective. This work has potential for technology that may transform the detection and treatment of diabetic foot ulcers and lead to a paradigm shift in the clinical care of the diabetic foot. This work has formed the basis to achieve future targets that include: 1) developing the automatic annotator that can automatically delineate and classify the foot images without the help of clinicians; 2) developing the automatic ulcer detection and recognition and segmentation with the help of these classifiers; 3) implementing the method to determine the various pathologies of DFU as multi-class classification similar to the Texas classification and other grading scales; 4) implementing the various user-friendly software tools including mobile applications for ulcer recognition \cite{yap2017footsnap}. Since DFUNet worked well for DFU classification, this proposed framework will likely be useful for classifying the other skin lesions such as wound classification, infections like chicken pox or shingles, other skin lesions like moles and freckles, spotting marks and pimples \cite{alarifi2017facial} against the normal skin. For classification, DFUNet is a light-weight CNN framework that is currently fine-tuned for only two classes (ulcer and normal skin), it will be further tested in the future to include many more classes. Therefore, we demonstrated how reducing the number of layers and fine-tuning using the bespoke architecture of DFUNet markedly reduced processing time, while also achieving a higher sensitivity and specificity.

%
%
%
%


%
\bibliographystyle{IEEEtran}
\bibliography{egbib}

%

\begin{IEEEbiography}[{\includegraphics[width=1in,height=1.25in,clip,keepaspectratio]{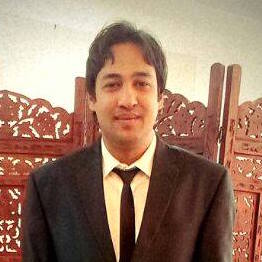}}]{Manu Goyal} is a Research Scholar at the Manchester Metropolitan University. He received his Master of Technology in Computer Science and Applications from Thapar University, India. His research expertise is in medical imaging analysis, computer vision, deep learning, wireless sensor networks and internet of things
\end{IEEEbiography}

\begin{IEEEbiography}[{\includegraphics[width=1in,height=1.25in,clip,keepaspectratio]{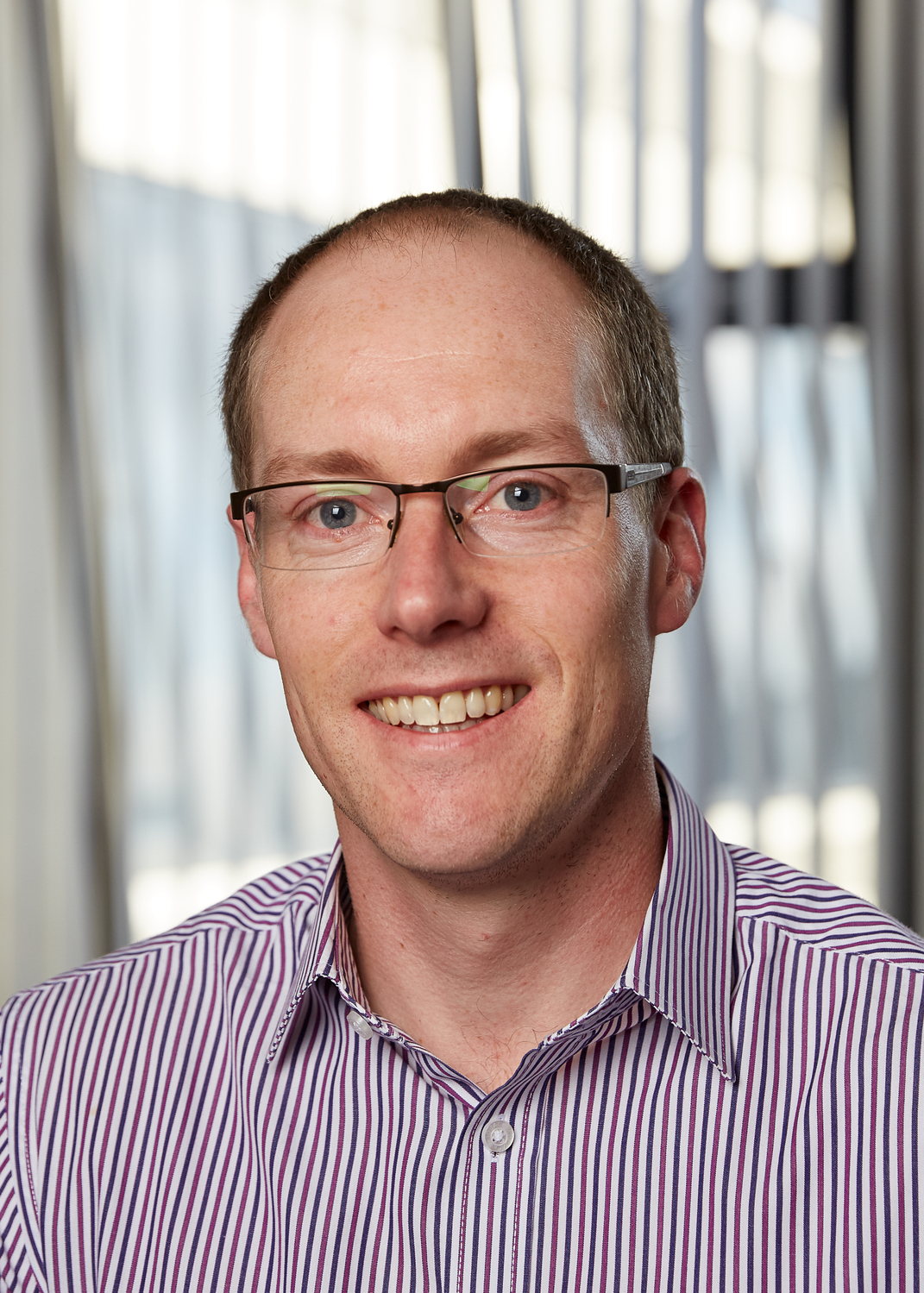}}]{Neil D. Reeves}
	is Professor of Musculoskeletal Biomechanics at the Manchester Metropolitan University. He conducts research into gait, movement impairment and foot pathology in diabetes and has published in journals such as Diabetes Care and Diabetic Medicine. He has led research projects on these topics supported by European and UK funding bodies.
\end{IEEEbiography}

\begin{IEEEbiography}[{\includegraphics[width=1in,height=1.25in,clip,keepaspectratio]{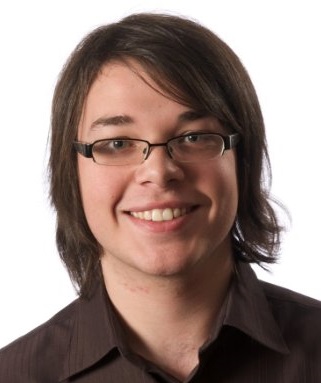}}]{Adrian K. Davison} is a Research Associate at the University of Manchester. He received his BSc (Hons.) degree in Multimedia Computing in 2012 and his PhD degree in 2016 from Manchester Metropolitan University. He had an active role as a student representative within MMU, Co-Chairing the internal MMU Science and Engineering Symposium 2015. His current role is the medical imaging analysis of children's hip disease. His research expertise is in facial expression and micro-facial expression analysis, deep learning and medical image analysis.
\end{IEEEbiography}

\begin{IEEEbiography}[{\includegraphics[width=1in,height=1.25in,clip,keepaspectratio]{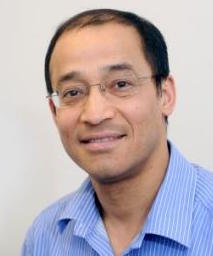}}]{Satyan Rajbhandari}
	 is a Consultant Physician and Honorary Clinical Professor at Lancashire Teaching Hospitals. He received his MD from Sheffield, FRCP from London and FRCP from Edinburgh. His specialist areas are Diabetes; endocrinology; general medicine. His research interests are Diabetes treatment; cardiovascular risks in diabetes; diabetic neuropathy; diabetic foot; epidemiology. He has various awards and prizes such as Arun Baksi Award for close co-operation between primary and secondary care diabetes service, Young Researcher Award, European Association of Study of Diabetic Neuropathy (2005), Sir Ernest Finch Travel Award for Young Researcher (2000), Award by Nepal Govt, 1988 for management of mass casualty, Colombo Plan Scholarship ( 1981-86), Mahendra Ratna Scholarship ( 1978-80), Highest scorer in national GCSE level (Nepal).
	
\end{IEEEbiography}

\begin{IEEEbiography}[{\includegraphics[width=1in,height=1.25in,clip,keepaspectratio]{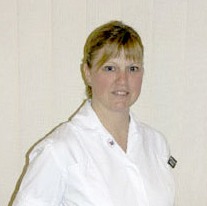}}]{Jennifer Spragg}
	 is a Podiatrist / Chiropodist in the Rossendale Practice, Rawtenstall, Lancashire. She graduated from University of Salford in 2005 and Registered with the Health Professions Council (HPC).She is also a member of the Society of Chiropodists and Podiatrists. She has 25 years experience as Registered General Nurse (RGN)- last 15 years as Practice Nurse in GP surgeries and has a wide range of knowledge of medical conditions and treatments.
\end{IEEEbiography}

\begin{IEEEbiography}[{\includegraphics[width=1in,height=1.25in,clip,keepaspectratio]{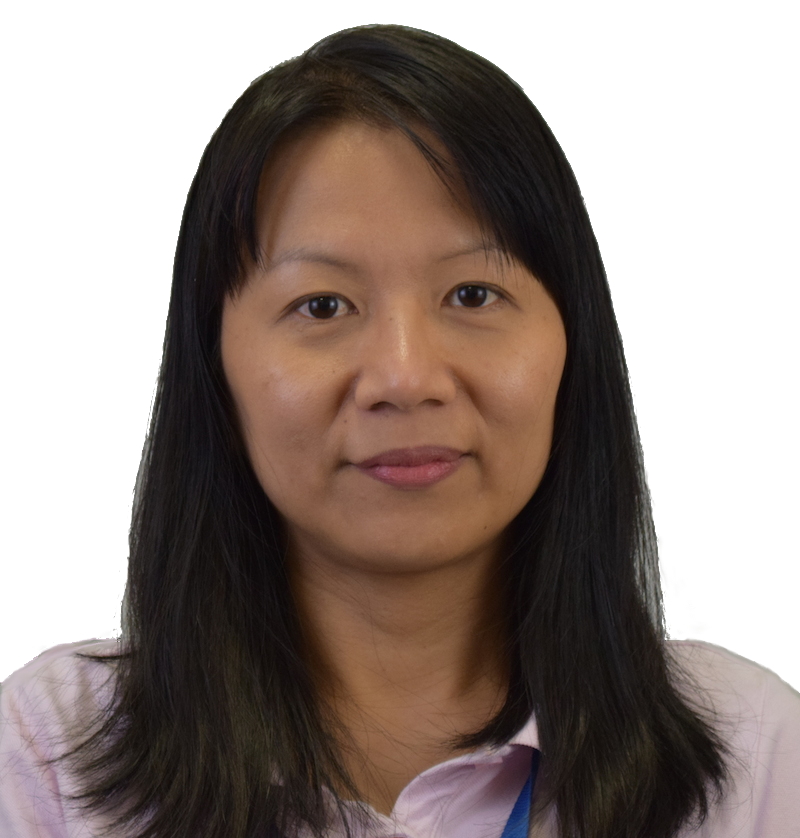}}]{Moi Hoon Yap} is a Reader (Associate Professor) in Computer Science at the Manchester Metropolitan University and a Royal Society Industry Fellow with Image Metrics Ltd.  She received her PhD in Computer Science from Loughborough University in 2009. After her PhD, she worked as Postdoctoral Research Assistant (April 09 - Oct 11) in the Centre for Visual Computing at the University of Bradford. She serves as an Associate Editor for Journal of Open Research Software and reviewers for IEEE transactions/journals (Image Processing, Multimedia, Cybernetics, biomedical health and informatics). Her research expertise is computer vision, applied machine learning and deep learning.
\end{IEEEbiography}

\end{document}